\documentclass[letterpaper, 10pt, conference]{ieeeconf}
\IEEEoverridecommandlockouts  % to override locked commands

\usepackage{cite}
\usepackage{amsmath,amssymb,amsfonts}
\usepackage{algorithmic}
\usepackage{graphicx}
\usepackage{textcomp}
\usepackage{xcolor}
\def\BibTeX{{\rm B\kern-.05em{\sc i\kern-.025em b}\kern-.08em
    T\kern-.1667em\lower.7ex\hbox{E}\kern-.125emX}}

\usepackage{multirow}
\usepackage{amsmath}
\usepackage[linesnumbered,ruled,vlined]{algorithm2e}
\usepackage{url}
\usepackage{booktabs}
\usepackage{threeparttable}

\begin{document}

\title{KD-GAT: Combining Knowledge Distillation and Graph Attention Transformer for a Controller Area Network Intrusion Detection System\\
}

\author{
Robert Frenken, Sidra Ghayour Bhatti, Hanqin Zhang, Qadeer Ahmed \\
The Ohio State University, Columbus, OH, USA \\
\{frenken.2, bhatti.39, zhang.14641, ahmed.358\}@osu.edu
}
% \author{
% \IEEEauthorblockN{Robert Frenken, Sidra Ghayour Bhatti, Hanqin Zhang, Qadeer Ahmed}
% \IEEEauthorblockA{The Ohio State University, Columbus, OH, USA \\
% \{frenken.2, bhatti.39, zhang.14641, ahmed.358\}@osu.edu}
% }
\maketitle
\renewcommand{\thefootnote}{}
\footnotetext{Code available at: \url{https://github.com/OSU-CAR-MSL/KD-GAT}}
\renewcommand{\thefootnote}{\arabic{footnote}}

\begin{abstract}
The Controller Area Network (CAN) protocol is widely adopted for in-vehicle communication but lacks inherent security mechanisms, making it vulnerable to cyber-attacks. This paper introduces KD-GAT, an intrusion detection framework that combines Graph Attention Networks (GATs) with knowledge distillation (KD) to enhance detection accuracy while reducing computational complexity. In our approach, CAN traffic is represented as graphs using a sliding window to capture temporal and relational patterns. A multi-layer GAT with jumping knowledge aggregation acting as the teacher model, while a compact student GAT—only 6.32\% the size of the teacher—is trained via a two-phase process involving supervised pretraining and knowledge distillation with both soft and hard label supervision. Experiments on three benchmark datasets—Car-Hacking, Car-Survival, and can-train-and-test—demonstrate that both teacher and student models achieve strong results, with the student model attaining 99.97\% and 99.31\% accuracy on Car-Hacking and Car-Survival, respectively. However, significant class imbalance in can-train-and-test has led to reduced performance for both models on this dataset. Addressing this imbalance remains an important direction for future work.
\end{abstract}

\section*{Keywords}
Controller Area Network, Intrusion Detection System, Cybersecurity, Graph Attention Network, Knowledge Distillation.
% \section*{Code Availability}
% Code available at: \url{https://github.com/OSU-CAR-MSL/KD-GAT}
% \begin{IEEEkeywords}
% Controller Area Network, Intrusion Detection System, Cybersecurity, Graph Attention Network, Knowledge Distillation.
% \end{IEEEkeywords}
% trim={<left> <lower> <right> <upper>}
% Put Figure 1 on top of page 2
\begin{figure*}[t]
\centering
\includegraphics[width=\textwidth, trim={0.0cm 5cm 0.0cm 5cm}, clip]{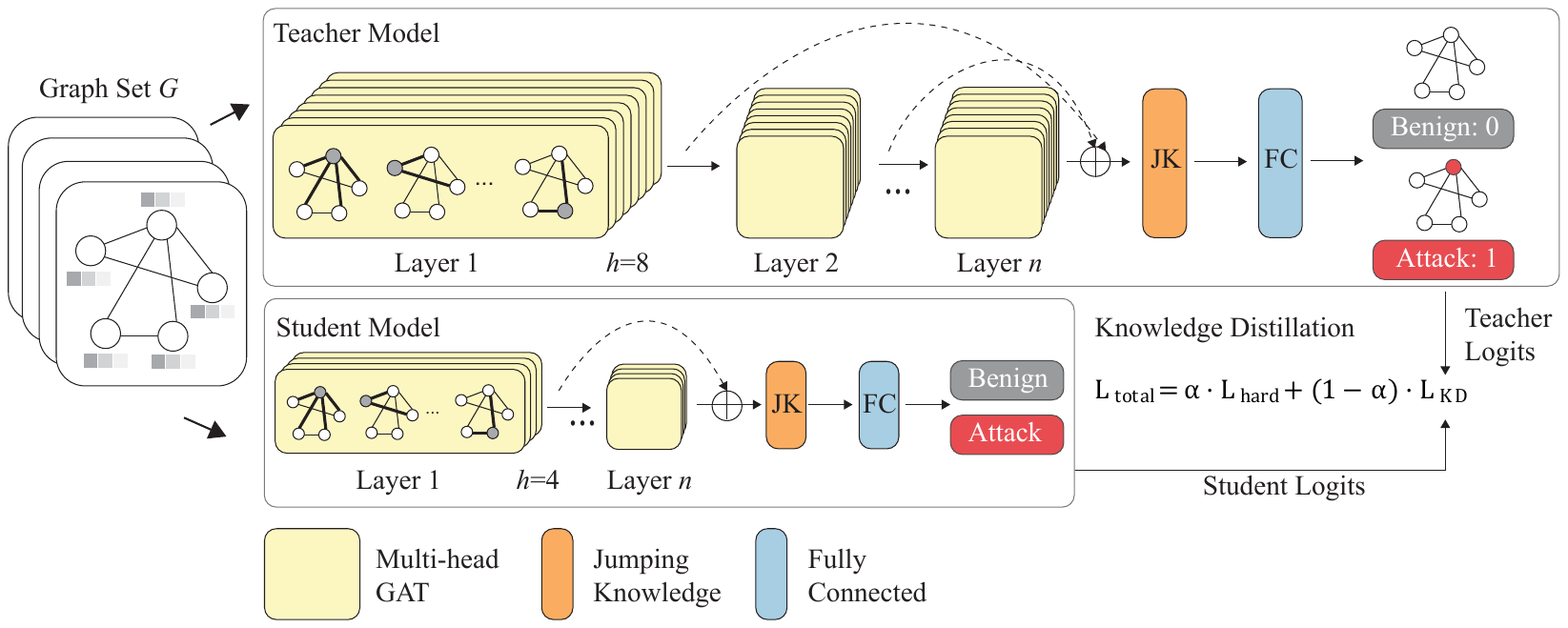}
\caption{KD-GAT knowledge distillation framework. Variables $h$ and $n$ denote the number of attention heads and number of layers.} 
%, while $t$ (teacher) and $s$ (student) denotes the hyperparameters for said variables.}
\label{fig:example}
\end{figure*}

\section{Introduction}
Modern vehicles contain numerous electronic control units (ECUs) that manage nearly all functions of the vehicle, ranging from engine performance to advanced driver assistance systems (ADAS), such as lane departure warning (LDW), and adaptive cruise control (ACC). The controller area network (CAN) protocol, introduced in the mid-1980s for in-vehicle network (IVN), is known for its reliability, cost-effectiveness, and robustness \cite{Pazul}. CAN is widely recognized as the standard communication protocol for ECUs in IVNs. However, the CAN protocol was originally developed with little emphasis on security features such as encryption, message authentication, and access control, as it was assumed to be a closed system without external access points \cite{ChoiK}.

With the adoption of the on-board diagnostic (OBD) port in all vehicles, direct access to the CAN network became possible. Access to the network has increased even further with the development of various communication technologies, such as Wi-Fi, vehicle-to-everything (V2X) networks, and cellular networks. This has allowed external sources, including other vehicles and infrastructure, to interact with the vehicle, thus expanding the potential attack surfaces of the vehicle CAN system \cite{Miller}. Users of open protocols like J1939, understanding this inherent vulnerability, have started developing security solutions, though there are still many cybersecurity gaps to close.

Studies \cite{Woo, Wen} have shown that cyberattacks on the in-vehicle CAN network can be launched via remote access points (e.g., Wi-Fi, mobile networks, Bluetooth) and physical interfaces (e.g., USB, OBD-II, CD players). Once access is gained, attackers can inject malicious messages to disrupt vehicle functions, potentially taking full control. Attacks on critical systems like brakes or powertrain can lead to severe safety risks, resulting in recent research on enhancing CAN bus security through intrusion detection systems.

With advances in AI, IVN intrusion detection has shifted from monitoring ECU physical characteristics to analyzing CAN data using deep learning \cite{Islam, Park}, primarily categorized into packet-based and window-based approaches.
% However, many supervised models label entire data flows as anomalous, lacking granularity in determining specific malicious messages, which limits detection accuracy and traceability. 
% To address rising CAN network threats, researchers have proposed various intrusion detection systems (IDS), primarily categorized into packet-based and window-based approaches.
Packet-based IDSs detect threats quickly using features like ECU clock skew and timing \cite{Cho}, but they can't capture packet sequences, limiting detection of spoofing, flooding, fuzzy, and replay attacks. Window-based IDSs analyze packet correlations, enabling detection and classification. 
Statistical methods like entropy \cite{Muter} struggle with low-volume attacks, while survival analysis \cite{Han} shows promise but lacks replay attack evaluation. 
To address limitations in window-based IDSs, a study by \cite {Islam} proposed a statistical method based on graph theory that builds one graph per N packets and uses a chi-square test to identify attack windows. 
More recent ML-based IDSs, such as the deep convolutional neural network (DCNN)-based IDS \cite{Kim} and the K-nearest neighbors (KNN)-based IDS \cite{Derhab}, aim to reduce such delays while improving detection accuracy.

In \cite{Malik}, an IDS using temporal graph-based features and machine learning (SVM, KNN) is proposed achieving up to 97.99\% accuracy against DoS, fuzzy, and spoofing attacks. 
A real-time CAN IDS using dynamic graphs is proposed to detect attacks and identify compromised message IDs efficiently in \cite{Jiaru}. It achieves low detection time and memory usage, with theoretical validation via probabilistic analysis. 
An A$\&$D-GNN-based IDS is proposed in \cite{AD} for CAN bus IDS using Arbitration and Data graphs. It achieves 99.92\% accuracy on the OTIDS dataset, outperforming existing methods.
Finally, KD-XVAE, a lightweight VAE-based IDS uses knowledge distillation and incorporates explainable AI via SHAP for transparent decision-making, ensuring robust and efficient IVN security \cite{KD}. ML-based IDSs are hard to adapt and deploy in resource-limited IVNs due to large models and low interpretability. To help address this, techniques such as knowledge distillation and utilizing graph-based features can enable lightweight and adaptable detection.

\subsection{Motivation}
Numerous IDS approaches have been studied for securing CAN protocol, but each technology has drawbacks. Packet-based IDSs
are limited in their ability to analyze the correlation of consecutive packets, and most are unable to classify attacks.
Window-based IDS have been studied as well, but development is still in its early stages. 
To address the limitations of existing IDSs, we propose a graph neural network (GNN)-based teacher model that can learn rich, structural patterns from the CAN traffic. Then, a lightweight student model is trained via knowledge distillation to replicate the teacher’s insights—enabling efficient and accurate intrusion detection even on resource-constrained in-vehicle systems. 
% ****add reference to some other pprs that how our model performs well****
\subsection{Contributions}
Our main contributions are as follows:
\begin{itemize}
    \item We propose a novel Graph Attention Transformer (GAT) architecture enhanced with jumping knowledge skip connections, enabling more effective learning of structural patterns in CAN traffic.
    \item We introduce a knowledge distillation framework tailored for resource-constrained edge environments, achieving a compact student model that is only 6.32\% the size of the teacher model while maintaining high detection performance.
    \item We conduct comprehensive experiments on multiple benchmark datasets, including the newly created can-train-and-test\cite{Lampe2024cantrainandtest}. To the best of our knowledge, this is the first deep learning study to evaluate the entire can-train-and-test dataset.
\end{itemize}

Figure \ref{fig:example} visualizes the framework.

\section{Related Work}
\subsection{Literature Review}
Several IDS techniques have been explored for securing in-vehicle CAN networks, with common classification based on three key aspects: the number of frames needed for detection, the type of data used, and the detection model design \cite{Dupont}. Based on frame count, IDSs are typically categorized into packet-based and window-based approaches.

Packet-based IDS approaches detect attacks using data from a single CAN packet. Kang et al. \cite{Kang} employed deep neural networks (NNs) to classify high-dimensional CAN packet features as normal or malicious, though their evaluation was limited to simulation data. Groza et al. \cite{Groza} developed an IDS leveraging the cyclic nature of in-vehicle traffic and fixed data field formats, using Bloom filters to assess frame periodicity. However, this method is only effective for periodic frames, not aperiodic ones \cite{Cheng}. 
While packet-based IDSs offer fast detection, they lack the ability to analyze correlations between consecutive packets, limiting their accuracy.

Window-based IDSs analyze CAN packet sequences within fixed sizes or time frames. Olufowobi et al. proposed a real-time IDS using timing models based on message intervals and worst-case response times. It detects anomalies without predefined specs but struggles with aperiodic messages and repeated IDs \cite{Olufowobi}. Several window-based IDSs use frequency analysis. Taylor et al. analyzed CAN packet frequency and Hamming distance \cite{Taylor}, but such methods struggle with aperiodic packet attacks \cite{Bozdal, Choi}. Islam et al. used graph features from fixed-size windows with chi-squared and median tests to detect anomalies and replay attacks, but their approach needs many packets, reducing responsiveness \cite{Islam}. 
Graph-based IVN IDSs model ECU interactions well but mainly detect simple attacks
G-IDCS addresses this by combining a threshold-based (TH) classifier for reduced window size and interpretability with an ML classifier that uses message correlations to classify attack types—something packet-based IDSs miss \cite{GIDCS}. 

\section{Background}
 \begin{figure}[h] 
    \centering
\includegraphics[width=0.48\textwidth, trim={0.0cm 0cm 0.0cm 0cm}, clip]{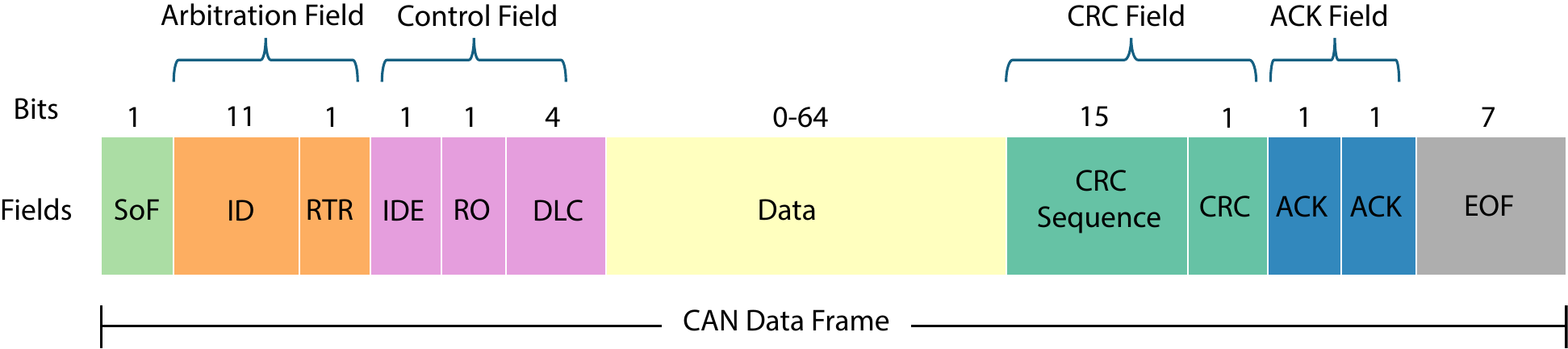}
   \caption{CAN data frame structure}
    \label{CAN_frame}
\end{figure}
\subsection{CAN Protocol}
The CAN is a serial communication system designed to efficiently handle distributed real-time control, and it is widely used for communication between electronic control units (ECUs) in vehicles. In a CAN bus setup, transmitting nodes broadcast messages, while receiving nodes use filtering mechanisms to decide whether to process the messages. As illustrated in Figure \ref{CAN_frame}, a CAN data frame is made up of seven fields. First is start-of-frame (SoF), a single dominant bit, then arbitration field, an 11-bit identifier and a 1-bit RTR. Next is the control field, a 1-bit IDE, a reserved bit, and a 4-bit DLC that specifies the data field’s length. The data field varies in size according to the DLC and holds the actual payload. The CRC field features a 15-bit cyclic redundancy check and a 1-bit delimiter with a recessive bit. The ACK field includes a dominant bit for acknowledgment and a recessive delimiter. Finally, the EoF field ends the frame with seven recessive bits.

\subsection{Graph Neural Networks}
A graph is a data structure consisting of a set of nodes  \textit{V} and a set of edges \textit{E} that connect pairs of nodes. A graph can be defined as $G = (V,E)$, where $V = \{v_1, v_2, ..., v_n\}$ is a node set with $n$ nodes, and $E = \{e_1, e_2, ..., e_m\}$ is an edge set with $m$ edges. 
%Edge $e_k = \{v_i, v_j\}$ indicates a connections between a pair of nodes. 
Given this graph structure, a graph neural network (GNN) looks to find meaningful relationships and insights of the graph. The most common way to accomplish this is through the message passing framework \cite{gilmer2017mpnn} \cite{scarselli2009gnn}, where at each iteration, every node aggregates information from its local neighborhood. Across iterations, node embeddings contain information from further parts of the graph. This update rule can be explained through the following equation:
% GNN Update Rule
\begin{equation}
\mathbf{h}_v^{(k)} = \phi\left(
    \mathbf{h}_v^{(k-1)},\ 
    \bigoplus_{u \in \mathcal{N}(v)} 
    \psi\left(\mathbf{h}_v^{(k-1)},\ \mathbf{h}_u^{(k-1)},\ \mathbf{e}_{vu}\right)
\right)
\end{equation}
where $\phi$ is the update function, $\psi$ is the message function, $\bigoplus$ is the aggregation operator (e.g., sum, mean), and $\mathcal{N}(v)$ is the neighbors of node $v$.

GAT \cite{velickovic2018gat} builds upon GNNs by introducing an attention mechanism. This allows each node in the message passing framework to dynamically assign weight contributions to their neighbors. For node $v$, the attention coefficient $\alpha_{vu}$ for neighbor $u$ is computed as:
% Graph Attention Transformer Equation
\begin{equation}
\alpha_{vu} = \mathrm{softmax}\left(
    \mathrm{LeakyReLU}\left(
        \mathbf{a}^\top 
        \left[
            \mathbf{W}\mathbf{h}_v \mathbin{\|} \mathbf{W}\mathbf{h}_u
        \right]
    \right)
\right)
\end{equation}
where $\alpha$ is the learnable attention vector, $\mathbf{W}$ is a weight matrix, and $\mathbin{||}$ denotes concatenation between the weight matrices.

% GAT Node Update
The attention function computes a scalar weight for each neighbor of node $v_i$, denoted by $\alpha_{ij}$, which reflects the importance or relevance of node $v_j$ for node $v_i$.
\begin{equation}
\mathbf{h}_v^{(k)} = \sigma\left(
    \sum_{u \in \mathcal{N}(v)} 
    \alpha_{vu} \mathbf{W} \mathbf{h}_u^{(k-1)}
\right)
\end{equation}
where $\sigma$ is the activation function, normally ELU or RELU.

The Jumping Knowledge (JK) module \cite{JumpingKnowledge} further enhances GATs by combining intermediate layer representations, similar to a skip connection. This paper uses the LSTM-based aggregation, where $h_v^{(l)}$ % \in \mathbb{R}^d$ 
denotes node $v$'s representation at layer $l \in {1,...,L}$. The JK module gathers all layer outputs, performs a forward LSTM pass through layers (1→L), and a backward LSTM pass through the same layers in reverse (L→1), where they are concatenated for a final representation. Equations \ref{LSTM_right}, \ref{LSTM_left}, and \ref{concat} demonstrate this process.
\begin{equation}
\label{LSTM_right}
    \overrightarrow{h}_v^{(l)} = \mathrm{LSTM}\left( h_v^{(l)}, \overrightarrow{h}_v^{(l-1)} \right)
\end{equation}
\begin{equation}
\label{LSTM_left}
    \overleftarrow{h}_v^{(l)} = \mathrm{LSTM}\left( h_v^{(l)}, \overleftarrow{h}_v^{(l+1)} \right)
\end{equation}
\begin{equation}
\label{concat}
    h_v^{\text{final}} = \left[ \overrightarrow{h}_v^{(l)} \,; \overleftarrow{h}_v^{(l)} \right]
\end{equation}

\section{Methodology} \label{methodology}

\subsection{Graph Construction}

CAN data is typically represented as a tabular time series dataset, where each message \( m_i \) is characterized by attributes such as CAN ID, payload data, and timestamp. To convert this sequential data into graph inputs suitable for graph-based intrusion detection, we define the following procedure:

\paragraph{Sliding Window and Graph Formation}  
Given a sliding window of size \( W \) (set to 50 messages in this work), we extract a subsequence of CAN messages:
\[
\mathcal{W}_t = \{ m_{t}, m_{t+1}, \ldots, m_{t+W-1} \}
\]
where each message \( m_i = ( \text{ID}_i, \text{payload}_i ) \).

\paragraph{Node Definition}  
Each unique CAN ID within the window \(\mathcal{W}_t\) corresponds to a node \( v_j \) in the graph \( G_t = (V_t, E_t) \). The node attributes are defined as:
\begin{equation}
\mathbf{x}_j = \left[
\text{ID}_j,
f_j = \frac{\text{count}(\text{ID}_j)}{W},
\bar{p}_j = \frac{1}{n_j} \sum_{k=1}^{n_j} \text{payload}_{j,k}
\right]
\end{equation}
where \( n_j \) is the number of occurrences of CAN ID \( j \) in the window, and \(\bar{p}_j\) is the average payload value for that ID.

\paragraph{Edge Construction}  
Edges \( e_{jk} \in E_t \) are created between nodes \( v_j \) and \( v_k \) if their corresponding messages appear sequentially in the window. Formally, for message pairs \((m_i, m_{i+1})\), if \( m_i \) corresponds to node \( v_j \) and \( m_{i+1} \) to node \( v_k \), then:
\[
e_{jk} = \text{number of occurrences of } (v_j, v_k) \text{ in } \mathcal{W}_t
\]
This captures the temporal relational structure of CAN messages within the window.

\begin{figure}[h]
        \centering
        \includegraphics[width=\linewidth, trim={0.5cm 0.20cm 0.5cm 0.60cm}, clip]{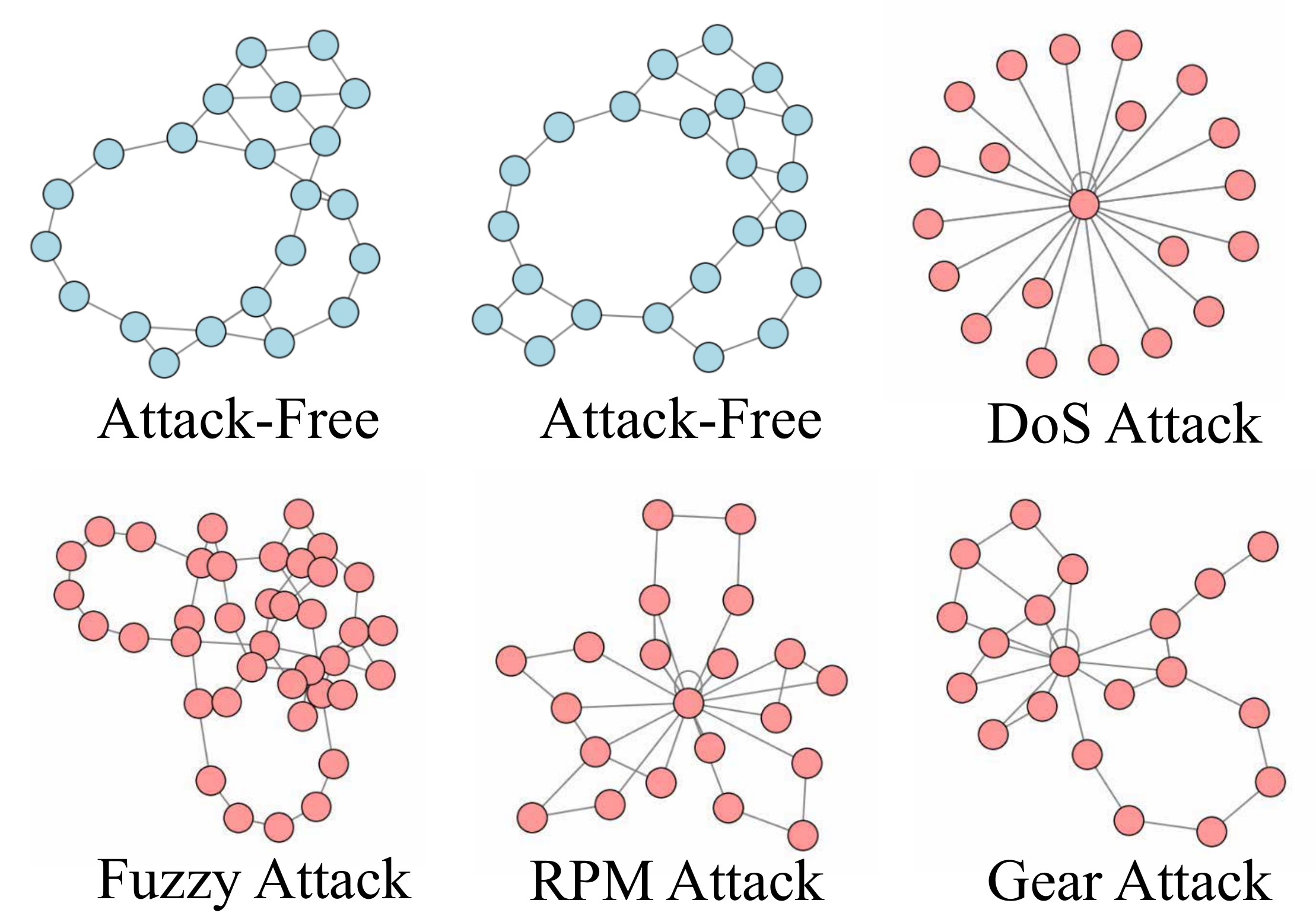}
        \caption{Graphs created from the Car Hacking Dataset. Each node represents a unique CAN ID found in the window, and each edge is constructed between sequential IDs, including self edges. The feature vector associated with each node is not shown. Blue denotes that the graph has no attack IDs, while red indicates that at least one of the nodes is an intrusion.}
        \label{fig:graph-examples}
    \end{figure}

\paragraph{Graph Input Summary}  
Thus, each sliding window \(\mathcal{W}_t\) is represented as a graph \( G_t = (V_t, E_t, X_t) \) where \(V_t\) are nodes (unique CAN IDs), \(E_t\) are edges (sequential message relations), and \(X_t\) are node attributes as defined above. Labels are created for a binary classification task, where $0$ if $W_t$ contains only benign messages, and $1$ if $W_t$ contains any attack messages. Figure \ref{fig:graph-examples} shows some examples of attack-free and attack graphs.

% trim={<left> <lower> <right> <upper>}

\subsection{Knowledge Distillation}

Knowledge distillation (KD), popularized by Hinton et al.\cite{hinton2015distilling}, is a widely adopted model compression technique where a small, efficient student model is trained to reproduce the behavior of a large, accurate teacher model. 
The soft target probabilities output by a teacher model encode rich relational information between classes that's often not captured by hard labels alone. Training a student model to match these softened outputs enables it to learn a more informative function approximation than training with one-hot labels alone.

Concretely, given an input $x$, the teacher produces a vector of logits $s^t(x)$, which are converted into a softened distribution via temperature scaling $\tau$:
\begin{equation}
\tilde{p}^t_k(x) = \frac{\exp(s^t_k(x)/\tau)}{\sum_j \exp(s^t_j(x)/\tau)}
\end{equation}

The student is trained to match these probabilities by minimizing the Kullback-Leibler divergence between teacher and student distributions (distillation loss), alongside the standard supervised classification loss:
\begin{equation}
\mathcal{L}_{\text{total}} = \alpha \cdot \mathcal{L}_{\text{hard}} + (1 - \alpha) \cdot \mathcal{L}_{\text{KD}}
\end{equation}

where $\alpha$ balances the contribution of ground truth and teacher supervision.

\section{Experiments}
\subsection{Experimental Setup}
All experiments were conducted using PyTorch and PyTorch Geometric. Model training and evaluation were performed on GPU clusters provided by the Ohio Supercomputer Center (OSC). Table \ref{tab:teacher_student} highlights the differences between the teacher and student model, where the student model is 6.32\% the size of the teacher model.
\begin{table}[h]
    \centering
    \caption{Teacher and Student Model Parameters}
    \begin{tabular}{lcc}
        \toprule
        \textbf{Parameter} & \textbf{Teacher} &\textbf{Student} \\ [2pt] % <-- Adds extra space here
        \midrule
        GAT layers      & 5      & 2     \\
        Attention heads   & 8      & 4      \\
        Hidden channels   & 32    & 32   \\
        Linear Layers   & 3    & 3   \\
        Loss function & BCE & KL Divergence \\
        % Learnable Parameters & 4,867k & 283k \\
        Total Parameters & 4,999,426 & 316,034 \\
        % Total parameters
        \bottomrule
    \end{tabular}
    \label{tab:teacher_student}
\end{table}
The KD framework operates where the teacher network trains first, followed by the student network which consists of two stages:
\begin{itemize}
    \item \textbf{Stage 1:} Pretrain the student model using binary cross-entropy (BCE) loss only, known as a "warm up" period.
    \item \textbf{Stage 2:} Fine-tune the student using a mixed loss, combining BCE with a distillation loss using logits from the teacher and student.
\end{itemize}
Table \ref{tab:hyperparams} details the hyper-parameters for this experiment. Finally, in each experiment, 80\% of the dataset was utilized for training, 20\% for validation, and a distinct test set compiled by the dataset providers was used.
\begin{table}[h]
    \centering
    \caption{Hyperparameters}
    \begin{tabular}{lcc}
        \toprule
        \textbf{Hyperparameter} & \textbf{Value} & \textbf{Notes} \\
        \midrule
        Learning rate        & 0.0005   & Adam optimizer \\
        Batch size           & 128     & - \\
        Epochs               & 100      & Saved best validation epoch \\
        Student warm-up epochs & 5     & Distillation only after warm-up \\
        Dropout & 0.2     & Across all GAT and linear layers \\
        Temperature scaling  & 2.0     & For soft targets \\
        $\alpha$ & 0.5 & Distillation coefficient \\
        Window size          & 50     & Sliding window length \\
        Stride size          & 50      & Step between windows \\
        \bottomrule
    \end{tabular}
    \label{tab:hyperparams}
\end{table}

\subsection{Datasets}

Our proposed method has been evaluated on three publicly available automotive CAN intrusion detection datasets, each offering distinct characteristics and challenges for comprehensive IDS evaluation. Below provides a detailed comparison of dataset specifications and characteristics.

\textbf{HCRL Car-Hacking:} This dataset contains CAN traffic from a Hyundai YF Sonata with four attack types: DoS, fuzzing, RPM spoofing, and gear spoofing. All attacks were conducted on a real vehicle, with data logged via the OBD-II port. The dataset includes 988,872 attack-free samples and approximately 16.6 million total samples across all attack types\cite{Song2020carhacking}.

\textbf{HCRL Survival Analysis:} Collected from three vehicles (Chevrolet Spark, Hyundai YF Sonata, Kia Soul), this dataset enables scenario-based evaluation with three attack types: flooding (DoS), fuzzing, and malfunction (spoofing). The dataset is structured with 627,264 training samples and four testing subsets designed to evaluate IDS performance across known/unknown vehicles and known/unknown attacks \cite{Han2018survival}.

\textbf{can-train-and-test:}\footnote{https://bitbucket.org/brooke-lampe/can-train-and-test-v1.5/src/master/} The largest dataset, containing CAN traffic from four vehicles across two manufacturers (GM and Subaru). It provides nine distinct attack scenarios including DoS, fuzzing, systematic, various spoofing attacks, standstill, and interval attacks. The dataset is organized into four vehicle sets (set\_01 to set\_04) with over 192 million total samples. Each set contains one training subset and four testing subsets following the known/unknown vehicle and attack paradigm\cite{Lampe2024cantrainandtest}.

The can-train-and-test dataset exhibits extreme class imbalance with attack-free to attack sample ratios ranging from 36:1 to 927:1 across different subsets. To address this challenge, we applied focal loss to both model objective functions, focusing learning on hard-to-classify minority class samples:

\begin{equation}
FL(p_t) = -(1-p_t)^{\gamma}\,\log(p_t)
\end{equation}

where $\gamma$ is the focusing parameter set to 1.0 in our experiments. This approach helps mitigate the impact of class imbalance on model performance, particularly for minority attack classes.

% \subsection{Evaluation Metrics}
% We report \textbf{accuracy} ($\frac{TP + TN}{TP + TN + FP + FN}$), \textbf{precision} ($\frac{TP}{TP + FP}$), \textbf{recall} ($\frac{TP}{TP + FN}$), and \textbf{F1-score} ($\frac{2 \times \text{Precision} \times \text{Recall}}{\text{Precision} + \text{Recall}}$), where $TP$, $TN$, $FP$, and $FN$ denote true/false positives/negatives.

\subsection{Evaluation Metrics}
We report accuracy, precision, recall, and F1-score, defined as follows: 
\begin{equation}
Accuracy = \frac{TP + TN}{TP + TN + FP + FN}
\end{equation}
\begin{equation}
Precision = \frac{TP}{TP + FP}
\end{equation}
\begin{equation}
Recall = \frac{TP}{TP + FN}
\end{equation} 
\begin{equation}    
F1-Score = \frac{2 \times \text{Precision} \times \text{Recall}}{\text{Precision} + \text{Recall}}
\end{equation} 
where TP, TN, FP, and FN denote true/false positives/negatives.

\section{Results and Discussion}
\subsection{Results}
Below are the results of the experiments described in previous sections. The comparison models used were A\&D GAT \cite{AD} and ECF-IDS\cite{li2024ecfids}, an IDS utilizing a cuckoo filter and a BERT model. These were picked as one is the only know GAT model for CAN IDS, while the other is the only known DL model that has evaluated the can-train-and-test dataset. The performance of the model was assessed using the validation set and held out test set provided by \cite{Lampe2024cantrainandtest}.
\begin{table}[h]
\centering
\caption{Validation and Test performance for all methods on CAN intrusion datasets.}
\setlength{\tabcolsep}{5pt}
\renewcommand{\arraystretch}{1.1}
\begin{tabular}{llcccc}
\toprule
\multicolumn{6}{c}{\textbf{Validation}} \\
\midrule
\textbf{Dataset} & \textbf{Method} & \textbf{Acc.} & \textbf{Prec.} & \textbf{Recall} & \textbf{F1} \\
\midrule
\multirow{3}{*}{Car Hacking}  
& A\&D\cite{AD}      & 0.9995 & 0.9994 & 0.9993 & 0.9994 \\
& Teacher            & 0.9995 & 1.0000 & 0.9983 & 0.9991 \\
& Student-KD         & 0.9995 & 1.0000 & 0.9982 & 0.9991 \\
\midrule
\multirow{2}{*}{Car Survival} 
& Teacher            & 1.0000 & 1.0000 & 1.0000 & 1.0000 \\
& Student-KD         & 1.0000 & 1.0000 & 1.0000 & 1.0000 \\
\midrule
\multirow{2}{*}{Set\_01} 
& Teacher            & 0.9981 & 1.0000 & 0.8908 & 0.9423 \\
& Student-KD         & 0.9991 & 1.0000 & 0.9478 & 0.9732 \\
\midrule
\multirow{2}{*}{Set\_02} 
& Teacher            & 0.9740 & 0.9980 & 0.6983 & 0.8216 \\
& Student-KD         & 0.9959 & 0.9970 & 0.9553 & 0.9757 \\
\midrule
\multirow{3}{*}{Set\_03} 
& ECF-IDS\cite{li2024ecfids} & 0.9996 & 0.9991 & 0.9984 & 0.9986 \\
& Teacher            & 0.9963 & 1.0000 & 0.8897 & 0.9416 \\
& Student-KD         & 0.9977 & 0.9976 & 0.9335 & 0.9645 \\
\midrule
\multirow{2}{*}{Set\_04} 
& Teacher            & 0.9977 & 1.0000 & 0.8509 & 0.9194 \\
& Student-KD         & 0.9975 & 1.0000 & 0.8362 & 0.9108 \\
\midrule
\multicolumn{6}{c}{\textbf{Test}} \\
\midrule
\textbf{Dataset} & \textbf{Method} & \textbf{Acc.} & \textbf{Prec.} & \textbf{Recall} & \textbf{F1} \\
\midrule
\multirow{2}{*}{Car Hacking}  
& Teacher            & 0.9977 & 0.9969 & 0.9985 & 0.9977 \\
& Student-KD         & 0.9997 & 1.0000 & 0.9993 & 0.9997 \\
\midrule
\multirow{2}{*}{Car Survival} 
& Teacher            & 0.9695 & 0.9415 & 0.9987 & 0.9692 \\
& Student-KD         & 0.9931 & 0.9897 & 0.9961 & 0.9929 \\
\midrule
\multirow{2}{*}{Set\_01} 
& Teacher            & 0.9923 & 1.0000 & 0.7706 & 0.8705 \\
& Student-KD         & 0.9929 & 0.9993 & 0.7874 & 0.8808 \\
\midrule
\multirow{2}{*}{Set\_02} 
& Teacher            & 0.9908 & 0.5446 & 0.2473 & 0.3402 \\
& Student-KD         & 0.9818 & 0.2034 & 0.3055 & 0.2442 \\
\midrule
\multirow{2}{*}{Set\_03} 
& Teacher            & 0.9698 & 1.0000 & 0.5811 & 0.7350 \\
& Student-KD         & 0.9824 & 1.0000 & 0.7554 & 0.8606 \\
\midrule
\multirow{2}{*}{Set\_04} 
& Teacher            & 0.8831 & 0.9999 & 0.4960 & 0.6631 \\
& Student-KD         & 0.8707 & 1.0000 & 0.4425 & 0.6135 \\
\bottomrule
\end{tabular}
\end{table}
\subsection{Discussion}
Looking at the results, both the teacher and student perform as well or better with comparable models for the Car Hacking and Car Survival datasets in both the validation and test sets. The results from can-train-and-test are also promising with the validation set. However, when testing on unseen test data, the large discrepancy of class ratios cause metrics such as precision, recall, and F1-score to under perform. The only paper outside the original authors who used dataset \cite{Lampe2024cantrainandtest} was \cite{li2024ecfids} with their model ECF-IDS. However, they only provide results for Set\_03, so it's unclear how other deep learning models perform across all subsets of can-train-and-test. Measures such as different dropout rates, class weight balancing, and focal loss  were tried to alleviate the class imbalance, but these experiments were not able to perform significantly better.

% Should I explain what UMAP is here?
% Figure \ref{fig:enter-label} uses UMAP, a dimension reduction technique to  visualize the features learned by the teacher KD-GAT model. It is observed that the features learned have good separability among the attack and attack free, though there are some attacks that the graph embeddings find similar to the attack free graphs. Future work is needed to better understand why some attack graphs have similar properties to attack free graphs.
% \begin{figure}[h]
%     \centering
%     \includegraphics[width=0.7\linewidth,height=0.2\textheight]{umap.png}
%     \caption{Node Embeddings after UMAP}
%     \label{fig:enter-label}
% \end{figure}
\section{Conclusion}
We presented KD-GAT, a framework that transforms CAN traffic into graph representations and applies knowledge distillation to develop a compact, accurate intrusion detection model suitable for edge devices. Our student model is just 6.32\% the size of the teacher, yet achieves 99.97\% and 99.31\% accuracy on the Car-Hacking and Car-Survival datasets, respectively. While both models performed well on validation data, severe class imbalance in the can-train-and-test dataset led to reduced test performance. Addressing this imbalance remains a key direction for future work.
% Sources are now in the file bib.bib. I will try to organize them
\bibliographystyle{ieeetr}
\bibliography{bib}
% \label{tab:related_work}
% \end{table}
% \begin{table}[ht]
% \centering
% \caption{Comparison of Related Work on Car-Hacking Dataset}
% \centering
% \begin{tabular}{llll}
% \toprule
% \textbf{Method} & \textbf{Model Type} & \textbf{Accuracy}  \\
% \midrule
% GGNB (2022)~\cite{ggnb2022} & Graph-based Gaussian NB & 99.61\% \\
% Gen. Classifier (2024)~\cite{anomaly2024} & VAE + Generative Classifier & N/A \\
% DL Models (2025)~\cite{dl2025} & LSTM / VGG-16 & LSTM: 99.89\%, VGG-16: 100\% \\
% \bottomrule
% \end{tabular}
% \label{tab:related_work}
% \end{table}

\end{document}